%% file: collas2024_conference.tex
\documentclass{article} 
\usepackage{collas2024_conference,times}
\usepackage{easyReview}
\usepackage{algorithm}
\usepackage{algpseudocode}
\usepackage{multirow}
\usepackage{amssymb}

\input{math_commands.tex}

\usepackage{hyperref}
\hypersetup{
    colorlinks=true,
    linkcolor=red,
    filecolor=magenta,
    urlcolor=blue,
    citecolor=purple,
    pdftitle={Overleaf Example},
    pdfpagemode=FullScreen,
    }

\usepackage{tikz}
\usepackage{pgfplots}

\title{Task agnostic continual learning with \\ Pairwise layer architecture}

\author{Santtu Keskinen \\
Unaffiliated \\
\texttt{santtu.keskinen@gmail.com}
}

\preprintcopy 

\begin{document}

\maketitle

\begin{abstract}
Most of the dominant approaches to continual learning are based on either memory replay, parameter isolation, or regularization techniques that require task boundaries to calculate task statistics. We propose a static architecture-based method that doesn't use any of these. We show that we can improve the continual learning performance by replacing the final layer of our networks with our pairwise interaction layer. The pairwise interaction layer uses sparse representations from a Winner-take-all style activation function to find the relevant correlations in the hidden layer representations. The networks using this architecture show competitive performance in MNIST and FashionMNIST-based continual image classification experiments. We demonstrate this in an online streaming continual learning setup where the learning system cannot access task labels or boundaries.
\end{abstract}

\section{Introduction}
The problem of catastrophic forgetting in neural networks is decades old \citep{MCCLOSKEY1989109} but remains essential. Sequential learning of tasks remains unpractical with a few exceptions, such as pretrain-finetune regime where the loss of performance on the old task is acceptable or reinforcement learning where input and output distribution shift is unavoidable but challenging \citep{zhang2018study}. Humans can learn countless tasks in sequence without a problem, which suggests something is lacking in how we construct or train neural networks. This inability to handle sequential tasks also challenges approaches like warm-starting neural network training \citep{ash2020warmstarting} or curriculum learning \citep{faber2023mnist}.

The most straightforward solution to continual learning is the replay of previous experiences from a replay buffer. Even naive implementations of memory rehearsal have been shown to be very effective in continual learning \citep{hsu2019reevaluating, prabhu2020gdumb}. However, relying on a replay buffer may lead the network to overfit to the stored samples and hurt generalization \citep{verwimp2021rehearsal}. In addition, most rehearsal algorithms work by storing exact copies of past input data, which is biologically implausible. Thus, we feel motivated to look for rehearsal-free solutions to sequential learning of tasks. Even if rehearsal-free methods perform worse than rehearsal methods, they might end up being important for some part of a lifelong learning agent, e.g., a short-term memory module that rapidly adapts to changes in the environment.

The benchmarks in continual learning typically have neatly divided tasks, and clear boundaries between the tasks, and algorithms designed to do well on these benchmarks use this task structure to their advantage. However, using the task information limits the general usability of the continual learning algorithm. We want to design algorithms for scenarios where task labels are unavailable, noisy, or misleading. As described in \citet{french_1992}, the challenge of continual learning is finding the optimal representational overlap between tasks. We want the algorithm to be task-agnostic and able to find this overlap on its own.

Sparse representations are key to finding and utilizing such overlaps. They enable the model to focus only on the key features relevant to the current task while minimizing interference with unrelated tasks. This is achieved by activating a limited set of neurons for each task, thus preserving crucial information and facilitating the learning of discriminative features \citep{french_1992, srivastava_2013, bricken2023sparse, aljundi2019selfless, lan2023elephant, shen2021algorithmic}. To make our hidden representations sparse, we use k-WTA with subtraction \citep{bricken2023sparse}.

We introduce the Pairwise interaction Layer (PW-layer) as an architecture-centric solution to catastrophic forgetting. PW-layer performs a type of feature crossing, where the sparse hidden representations are expanded into all possible pairwise products (or crosses) of the said features, a.k.a. cross features. Since the full pairwise expansion would be extremely costly to compute for wide hidden layers, we apply extreme parameter sparsity after the pairwise expansion. The parameter sparsity gives us granular control on the number of trainable weights and how much total compute the PW-layer uses. We replaced our networks' final fully connected layer with a PW-layer and showed that this improves performance on split MNIST, permuted MNIST, and split FashionMNIST.

To test the new architectures in a task-agnostic way, we try two algorithms for computing the importance of parameters: Adagrad \citep{adagrad} and MAS \citep{aljundi2018memory} based Streaming Memory Aware Synapses (S-MAS). We use these algorithms to adjust the learning rate of the model's parameters, enabling continuous learning without explicit task boundaries.

The main contributions of our work are:

$1)$ Introduction of the Pairwise interaction layer for continual learning and experimental results showing its effectiveness in rehearsal-free continual learning.

$2)$ The evaluation of Adagrad and the introduction of S-MAS for online streaming calculation of parameter importance, highlighting their utility in continual learning without the need for explicit task structure.

$3)$ Experiments that not only show the benefits of our novel ideas but also show that a regular fully connected network with k-WTA sparsity is a solid rehearsal-free baseline in both Split and Permuted MNIST.

\section{Background and Related Work}

Some of the earlier attempts at tackling catastrophic forgetting were made in \citet{french_1992} with semi-distributed or sharpened representations, and \citet{srivastava_2013} who found that "local competition" in the hidden representations helps networks forget less. These early works showed that network architecture and sparsity matter in continual learning. However, we find that, with few exceptions (e.g., \citet{mirzadeh2022architecture, lan2023elephant}), work on continual learning places more emphasis on the learning algorithm itself rather than the network architecture. \citet{goodfellow2015empirical} went so far as to criticize the earlier work on LWTA and wrote: "In our more extensive experiments, we found that the choice of activation function has a less consistent effect than the choice of training algorithm ... We also reject the idea that hard LWTA is particularly resistant to catastrophic forgetting in general".

Since then, the field has seen a lot of progress, the vast majority of which has focused on things other than the core network architecture:

\textbf{Regularization methods} such as EWC \citep{Kirkpatrick_2017}, SI \citep{zenke2017continual}, MAS \citep{aljundi2018memory} try to limit the change of parameters in a way that the network can still solve old tasks by augmenting the loss function with regularization terms. \textbf{Parameter isolation methods} like HAT \citep{hat_serrà2018overcoming} and PackNet \citep{mallya2018packnet} train only a subset of weights for each task, whereas \textbf{Dynamic architecture methods} like Progressive neural networks \citep{rusu2022progressive} and VariGrow \citep{varigrow} grow the network for each task and freeze some of the parts trained on old tasks. \textbf{Gradient constraint methods} like Orthogonal Gradient Descent \citep{farajtabar2019orthogonal} and GPM \citep{saha2021gradient} set limits on how much and in which directions the parameters are allowed to be changed during training.

None of these works are particularly interested in the architecture of the trained network, and all use algorithms that perform some special operation when the task is switched and thus cannot work in a task-agnostic manner. Some of them could be made to work without task boundaries, but this is usually not straightforward and would presumably hurt their performance. Note that we don't count efforts like Task-free MAS \citep{aljundi2019taskfree} or Online EWC \citep{schwarz2018progress} as truly task-agnostic since both of them try to infer task boundaries from the data, which assumes that a task structure exists in the input stream. In contrast, our Streaming-MAS works completely without task boundaries and, when used with sparse representations, works about as well as the original MAS with task boundaries.

One class of continual learning methods that typically can be made task agnostic is based on \textbf{Variational inferece} or \textbf{Bayesian networks}. For example, UCB \citep{ebrahimi2020uncertaintyguided} uses the uncertainty inherent in Bayesian networks to calculate learning rates for each parameter in a task-agnostic way. We note that this approach is rather similar to ours on a high level. However, the authors of UCB provided only partial, non-working code, and we could not reproduce good results with UCB (the authors of VariGrow were also unable to reproduce the results of the UCB paper \citep{varigrow}). UCL \citep{NEURIPS2019_2c3ddf4b} is another similar variational inference-based method that works well in the multi-head Split MNIST but fails to generalize to the harder single-head Split MNIST task. We tried to include aspects of variational inference and uncertainty regularization in this study but cut them from the final version because they performed worse than our baseline, k-WTA, with Adagrad.

Recently \citet{zajac2024prediction} proposed a very successful generative method for continual learning of classifiers. Our method does not produce as good results as the proposed generative methods, but we significantly improve on the discriminative model baselines presented in the study. The generative method is somewhat limited in that they train separate generative networks for each class which requires class labels for the training. In comparison the methods described in this paper can work with any loss function and only trains a single model, even though it is only tested on classification tasks in this paper.

Most significant inspirations for this study come from SDMLP \citet{bricken2023sparse} and Elephant networks \citet{lan2023elephant}, which are two recent works on usage of \textbf{Sparse representations} in continual learning. However, our experimental results are better, and we also have removed some of the components that these papers claimed were essential for their success, e.g., we don't anneal the sparsity like SDMLP does or use sparse gradients like the elephant networks. This removal of concepts is necessary for simplicity and advancing the understanding of sparse activation functions in continual learning.

I previously compared K-WTA activation against other sparse activation functions in \citet{keskinen2024hard} and found that Hard ASH activation performs slightly better than K-WTA with regular fully connected network. However in our experiments K-WTA worked better with the pairwise architecture.

\section{Proposed Approach}

\colorlet{mylightred}{red!95!black!30}
\colorlet{mylightblue}{blue!95!black!30}
\colorlet{mylightgreen}{green!95!black!30}
\tikzset{ 
  >=latex, 
  node/.style={thick,circle,draw=#1!50!black,fill=#1,
                 minimum size=\pgfkeysvalueof{/tikz/node size},inner sep=0.5,outer sep=0},
  node/.default=mylightblue, 
  tconnect/.style={thick,blue!80!black!35,->}, 
  sconnect/.style={dashed,black!35}, 
}
\begin{figure}[h]
    \centering
    \begin{tikzpicture}[scale=1.2, every node/.style={scale=0.8}]
        \begin{scope}[shift={(0,0)}]
            \foreach \i in {1,...,4}
                \node[node, fill=mylightgreen, minimum size=0.75cm] (In\i) at (0,-\i) {$x_{\i}$};
            \foreach \i in {1,...,3}
                \node[node, fill=mylightred, minimum size=0.75cm] (Out\i) at (2,-\i-0.5) {$y_{\i}$};
            \foreach \i in {1,...,4}
                \foreach \j in {1,...,3}
                    \draw[tconnect] (In\i) -- (Out\j);
            \node at (1,0) {(a)};
        \end{scope}
        
        \begin{scope}[shift={(3.7,0)}]
            \foreach \i in {1,...,4}
                \node[node, fill=mylightgreen, minimum size=0.75cm] (BIn\i) at (0,-\i) {$x_{\i}$};
            \node[node, fill=mylightblue, minimum size=0.5cm] (BPair1) at (1.5,{-(1-0.3)*0.75}) {$x_1x_2$};
            \node[node, fill=mylightblue, minimum size=0.5cm] (BPair2) at (1.5,{-(2-0.3)*0.75}) {$x_1x_3$};
            \node[node, fill=mylightblue, minimum size=0.5cm] (BPair3) at (1.5,{-(3-0.3)*0.75}) {$x_1x_4$};
            \node[node, fill=mylightblue, minimum size=0.5cm] (BPair4) at (1.5,{-(4-0.3)*0.75}) {$x_2x_3$};
            \node[node, fill=mylightblue, minimum size=0.5cm] (BPair5) at (1.5,{-(5-0.3)*0.75}) {$x_2x_4$};
            \node[node, fill=mylightblue, minimum size=0.5cm] (BPair6) at (1.5,{-(6-0.3)*0.75}) {$x_3x_4$};
            \foreach \i in {1,...,3}
                \node[node, fill=mylightred, minimum size=0.75cm] (BOut\i) at (3,-\i-0.5) {$y_{\i}$};
            \draw[sconnect] (BIn1) -- (BPair1);
            \draw[sconnect] (BIn1) -- (BPair2);
            \draw[sconnect] (BIn1) -- (BPair3);
            \draw[sconnect] (BIn2) -- (BPair1);
            \draw[sconnect] (BIn2) -- (BPair4);
            \draw[sconnect] (BIn2) -- (BPair5);
            \draw[sconnect] (BIn3) -- (BPair2);
            \draw[sconnect] (BIn3) -- (BPair4);
            \draw[sconnect] (BIn3) -- (BPair6);
            \draw[sconnect] (BIn4) -- (BPair3);
            \draw[sconnect] (BIn4) -- (BPair5);
            \draw[sconnect] (BIn4) -- (BPair6);
            \foreach \i in {1,...,6}
                \foreach \j in {1,...,3}
                    \draw[tconnect] (BPair\i) -- (BOut\j);
            \node at (1.5,0) {(b)};
        \end{scope}
        
        \begin{scope}[shift={(8,0)}]
            \foreach \i in {1,...,4}
                \node[node, fill=mylightgreen, minimum size=0.75cm] (CIn\i) at (0,-\i) {$x_{\i}$};
            \node[node, fill=mylightblue, minimum size=0.5cm] (CPair1) at (1.5,{-(1-0.3)*0.75}) {$x_1x_2$};
            \node[node, fill=gray!50, minimum size=0.5cm] (CPair2) at (1.5,{-(2-0.3)*0.75}) {$x_1x_3$};
            \node[node, fill=mylightblue, minimum size=0.5cm] (CPair3) at (1.5,{-(3-0.3)*0.75}) {$x_1x_4$};
            \node[node, fill=gray!50, minimum size=0.5cm] (CPair4) at (1.5,{-(4-0.3)*0.75}) {$x_2x_3$};
            \node[node, fill=gray!50, minimum size=0.5cm] (CPair5) at (1.5,{-(5-0.3)*0.75}) {$x_2x_4$};
            \node[node, fill=mylightblue, minimum size=0.5cm] (CPair6) at (1.5,{-(6-0.3)*0.75}) {$x_3x_4$};
            \foreach \i in {1,...,3}
                \node[node, fill=mylightred, minimum size=0.75cm] (COut\i) at (3,-\i-0.5) {$y_{\i}$};
            \draw[sconnect] (CIn1) -- (CPair1);
            \draw[sconnect] (CIn1) -- (CPair3);
            \draw[sconnect] (CIn2) -- (CPair1);
            \draw[sconnect] (CIn3) -- (CPair6);
            \draw[sconnect] (CIn4) -- (CPair3);
            \draw[sconnect] (CIn4) -- (CPair6);
        
            \draw[tconnect] (CPair1) -- (COut1);
            \draw[tconnect] (CPair3) -- (COut3);
            \draw[tconnect] (CPair6) -- (COut2);
            
            \node at (1.5,0) {(c)};
        \end{scope}
    
    \end{tikzpicture}
    \caption{Illustration of (a) a normal fully connected layer with 4 inputs and 3 outputs, (b) a fully connected pairwise layer with 4 inputs, 6 expanded pairwise feature cross nodes and 3 outputs, and (c) a sparse pairwise interaction layer with just 3 trainable weights. Solid lines represent trainable weights. Each feature cross node multiplies 2 of the inputs together (illustrated by the dashed lines). The grey feature cross nodes are not connected to any outputs and can be pruned to save compute.}
    \label{fig:pairwise}
\end{figure}
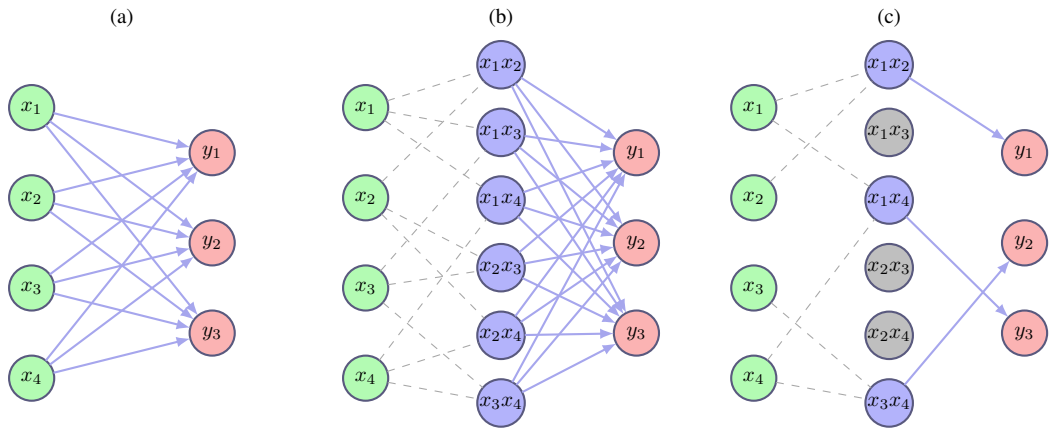

Our proposed task-agnostic continual learner has three key components:

$1)$ Sparse hidden layer activations from k-WTA type activation function.

$2)$ Pairwise interaction layer that expands the sparse activations into a higher-dimensional space of pairwise feature crosses, where each pairwise node represents the interaction between pairs of neurons from the second to last layer.

$3)$ Streaming continual learning algorithm that updates the per-parameter importance values that are used to adapt the learning rates for those parameters.

\subsection{Sparse activations with k-WTA}

For the sparse activation, we use k-WTA (also known as Top-k) with subtraction defined in \citet{bricken2023sparse}. In this activation, the $(k + 1)$th highest activation is first subtracted from all the hidden layer activations. Then, a ReLU is applied, leaving only the top-k values higher than the subtracted value.

Sparse activation functions have been used repeatedly in continual learning, e.g., in \citet{srivastava_2013, bricken2023sparse, lan2023elephant, Iyer_2022}. However, we feel that none of these works fully showcase how effectively an activation function like k-WTA reduces catastrophic forgetting. This is likely because each recent work uses a more complicated algorithm that obfuscates how much of the benchmark performance comes from the choice of activation function alone. In sections \ref{sec:split-mnist} and \ref{sec:sgd}, we show how even our baseline results with a regular fully connected layer score higher than popular regularization methods from the literature. This success of our fully connected baselines can be largely attributed to the k-WTA activation before the final layer.

\subsection{Pairwise Interaction Layer}
In the pairwise interaction layer (PW-layer), we leverage the concept of feature crossing. The basic idea is that the combination of features can provide more discriminative power than the individual features alone \citep{rendle_factorization}. Figure \ref{fig:pairwise}b depicts how the expansion is done. Every possible combination of two input neurons is multiplied together and forms a new cross feature. Only these expanded cross features are connected to the layer outputs. Intuitively, the pairwise crossing is a sensible operation to apply after k-WTA-like activation; for a pairwise cross node to be active, both pairwise inputs need to be in the set of $k$ most active inputs. This makes the PW-layer a type of filter that looks for patterns of pairs of inputs that are highly active together.

Feature transformations like crossing are not typical in modern neural network architectures, which rely on deep layers to automatically learn representations and feature interactions. However, explicitly modeling feature interactions, such as through feature crossing, can significantly enhance a model's ability to capture complex patterns and relationships in the data in shallow and wide networks \citep{cheng2016wide}. We argue that engineered feature transformations are useful in continual learning, where deep networks are more challenging to train \citep{lesort2022continual, lan2023elephant}.

Another feature of the PW-layer is the parameter sparsity. Since the total number of pairwise cross features is $\frac{d(d-1)}{2}$, where d is the width of the input to the PW-layer, it quickly becomes unpractical to densely connect all of the cross features to all outputs. For example, a fully connected PW-layer with an input width of 3000 and 100 output classes would have 450 million trainable weights. Therefore, instead of connecting all the cross features to all the output classes, we simply specify the number of trainable parameters we want the layer to have and randomly pick the included connections. In practice, our PW-layers have so few trainable parameters that we connect each cross-feature to either 1 or 0 outputs. Figure \ref{fig:pairwise}c shows an example of our sparsity scheme.

In our experiments, we only consider using a PW-layer just before the network's final output layer. We also tried to stack two PW-layers on top of each other but found that setup rather tricky to train. This might be because the PW-layers have a sparsity amplifying effect, and two of them together might limit the gradient flow too much.

\subsection{Streaming Continual Learning}

\noindent\begin{minipage}{\textwidth}
  \begin{minipage}[t]{0.45\textwidth}

    Our simple streaming continual learning algorithm \ref{alg:streaming} is designed to address the challenge of learning from a non-i.i.d. data stream without the need for explicit task boundaries. This approach focuses on dynamically adjusting the learning rates of the network parameters based on their importance to previously learned inputs, thereby balancing the trade-off between stability (the ability to retain old knowledge) and plasticity (the ability to learn new information). The importance of learning rate and adaptive learning rates in continual learning has been studied before \citep{mirzadeh2020understanding, ebrahimi2020uncertaintyguided, Tseran2018NaturalVC}, but we feel our approach is very natural and straightforward.

  \end{minipage}%
  \hfill 
  \begin{minipage}[t]{0.5\textwidth}
    \vspace{-0.8cm}
    \begin{algorithm}[H]
      \caption{Streaming Continual Learning}
      \label{alg:streaming}
      \begin{algorithmic}[1] 
        \State \textbf{Given:}
        \State \hspace{\algorithmicindent} Non-i.i.d. data stream
        \State \hspace{\algorithmicindent} Network weights $\theta$, randomly initialized
        \State \hspace{\algorithmicindent} Parameter importance $\Omega$, initialized with $\Omega \gets 0$
        \State \hspace{\algorithmicindent} Learning rate $\eta$ and a small constant $\epsilon = 1\times10^{-6}$
        \State \hspace{\algorithmicindent} Adagrad or S-MAS $UpdateRule$
        \State \hspace{\algorithmicindent} Constant $\lambda$, which controls stability-plasticity
        \For{$(x, y)$ from data stream}
            \State Compute $\nabla_\theta \mathcal{L}$ for the batch $(x, y)$

            \State $\Omega \gets$ $\Omega + \lambda \cdot UpdateRule(x, y, \theta, \nabla_\theta \mathcal{L})$
            \State $\theta \leftarrow \theta - \eta \cdot \nabla_\theta \mathcal{L} \cdot (\Omega + \epsilon)^{-\frac{1}{2}}$
        \EndFor
      \end{algorithmic}
    \end{algorithm}
  \end{minipage}
\end{minipage}

The basic premise of the algorithm is to maintain a measure of importance for each parameter in the network. This importance measure monotonically increases and is updated continuously with a simple rule (either Adagrad or S-MAS) as the network encounters new data. The learning rate for each parameter is then adjusted to be inversely proportional to the square root of its importance, allowing the network to modify less important parameters more freely while preserving the knowledge encoded in more critical parameters.

The hyperparameter $\lambda$ allows for fine-tuning the balance between retaining old knowledge (high $\lambda$) and acquiring new information (low $\lambda$), addressing the stability-plasticity dilemma inherent in continual learning.

Two primary rules for updating the parameter importance values are considered: Adagrad and Streaming Memory Aware Synapses (S-MAS).

\textbf{Adagrad} \citep{adagrad} is a well-known optimization algorithm that adapts the learning rates of all parameters by scaling them inversely proportional to the square root of the sum of all their past squared gradients. The update rule for Adagrad is simply: $UpdateRuleAdagrad(\nabla_\theta \mathcal{L}) = (\nabla_\theta \mathcal{L}) ^ 2$. Setting $\lambda = 1$ returns regular Adagrad, but we used $\lambda = 0.8$ for all our experiments, making our version of Adagrad favor plasticity a little more than regular Adagrad. I previously studied the effectiveness of Adagrad for continual learning in \citet{keskinen2024hard} where it performed the best out of popular optimizers.

\textbf{S-MAS} is based on the importance calculation from the Memory Aware Synapses (MAS) \citep{aljundi2018memory} algorithm, originally designed for continual learning with clear task boundaries. Unlike Adagrad, which uses gradient magnitude as the basis for importance, S-MAS directly assesses the sensitivity of the learned function to changes in each parameter, aiming to capture a more direct measure of each parameter's contribution to the current network output. The update rule for S-MAS is as follows:
\begin{center}
    $UpdateRuleSMAS(x, \theta) = \left| \nabla_\theta \left( \frac{1}{N} \sum_{i=1}^{N} ||f_\theta(x_i)||^2 \right) \right|$
\end{center}
Where $f_\theta$ is the neural network function and $N$ is the number of samples in a mini-batch.

The S-MAS is more expensive computationally than Adagrad since we always have to do two backward passes through the network: One to get the gradient of the loss and a second one to get the gradient of the $l_2$-normed learned function output.

Unlike in the original MAS, we do not need to store any old versions of the network parameters; the importance measure is only used to adapt the per-parameter learning rates, and there are no auxiliary losses. The importance is also updated online, whereas, in MAS, it is computed at the end of each task with a replay of the task data. Despite these simplifications, S-MAS seems to perform about as well as MAS, at least with the architectures tested in this study. However, it must be said that we did not test MAS as extensively as S-MAS.

With S-MAS, we used $\lambda = 0.01$ in all of the MNIST benchmarks, except the batch size experiments in section \ref{sec:batch_size}, and $\lambda = 0.001$ for the CIFAR-10 experiments in the appendix \ref{sec:cifar}

\section{Experiments and Results}
\subsection{Experiment Setup}

We tested the Pairwise output layer, Adagrad, and S-MAS using different network architectures. The experiments were conducted on Split MNIST \citep{lecun_1998, srivastava_2013}, Permuted MNIST \citep{Kirkpatrick_2017} and Split Fashion-MNIST \citep{xiao2017fashionmnist}.

In all of our experiments, the training algorithm is given the tasks for that experiment one by one in a sequence with no replay of previous inputs. Additionally, in the spirit of online learning, the network sees each input data point only once, i.e., we only train the network for a single epoch.

We did not do exhaustive hyperparameter searches. Instead, we did a few smaller trial runs to find an acceptable learning rate and amount of WTA sparsity for each architecture and experiment. Batch size was 64 for all experiments except those in \ref{sec:batch_size}. We did not tune $\lambda$ beyond finding the first value that seems to work well. This certainly means that some results could be improved further, but we tried to be fair between the baseline k-WTA FC architectures and the headline pairwise models.

\subsubsection{Split experiment setup}

In Split MNIST and Split Fashion-MNIST, the data was segmented into five subsets containing two classes each. This setup aims to simulate a continual learning environment where the model sequentially learns to classify different sets of classes. It's a setup that tests the model's ability to adapt to new tasks without revisiting previously encountered data,

We did all the split experiments for 30 different random network initializations and shuffled the task order between runs. We report the mean validation accuracies for those 30 runs. All results have standard errors under 1\% with a mean standard error of 0.4\%. The shuffling of the task order leads to quite a bit of variation in the results, but we feel it's important in order to make it harder to finetune the hyperparameters to the experiment.

For each split experiment we report two different results, single-head result and multi-head result. This distinction has a few names in the literature, such as Class-incremental vs. Task/Domain-incremental, but these terms can sometimes get muddled when researchers use various tricks to leak task information to the model, such as the "labels trick" \citep{zeno2019task}, special losses like in \citet{rebuffi2017icarl} or do parameter isolation based on task labels while still technically doing "incremental class learning" like in \citet{hat_serrà2018overcoming}. Therefore, we feel it is important to be very precise about what we mean by single-head and multi-head. In the single-head cases, neither the network being trained nor the training algorithm itself knows which task is currently trained, and the loss is always a generic softmax loss over all the possible output classes. In the multi-head setup, the network still doesn't get to use a task ID in the classification, but the loss and evaluation functions only consider the options that are possible for the current task.

\subsubsection{Experiment network architectures}

The experiments employed a variety of Multi-Layer Perceptrons (MLPs) and Convolutional Neural Networks (CNNs). For the MLPs, we narrow the hidden layer widths with the pairwise outputs to ensure that there is a roughly equivalent number of parameters across different configurations. We do this because it is well known that network width generally helps in continual learning \citep{mirzadeh2022wide}, and this strategy ensures that the comparisons between models focus on architectural differences and the impact of the Pairwise Layer rather than on variations in model size. For the CNNs, we do not try to equalize the parameter counts because the convolutions use only very few parameters, which makes balancing the parameter counts difficult, but we just present the results as they are.

Inside the network backbones, we always use the GELU activations \citep{hendrycks2023gaussian}, followed by k-WTA before the final layer, either a pairwise or fully connected layer. Interestingly enough, we found that adding another GELU just before the k-WTA improved the final accuracy by a percent or so.

For example, a network with MLP 3 x 700 (3 times 700 hidden neurons) backbone and Pairwise output has the following layers: 3xDense-GELU)-WTA-Pairwise.

The two CNNs follow a simple pattern: the first layer is a 7x7 convolution with the stride of 4, and the optional second layer is a 5x5 convolution with the stride of 2. The convolution layers are likewise always followed by a GELU. There are no pooling or normalization layers.

\subsection{Performance on Split MNIST}
\label{sec:split-mnist}
\begin{table}[h]
\centering
\caption{Split-mnist}
\label{tab:split-mnist}
\begin{tabular}{ccc|cc|cc}
Network &
Output &
Total &
Single-head &
Single-head &
Multi-head &
Multi-head \\ 
backbone & head type & parameters & Adagrad & S-MAS & Adagrad & S-MAS
\\ \hline
MLP 1 x 700   & Pairwise & 0.8M           & 83.1 & \textbf{84.4} & \textbf{99.6} & 98.8 \\
MLP 1 x 1000  & k-WTA FC & 0.8M           & 72.9 & 72.6          & 99.2          & 98.1 \\ \hline
MLP 1 x 3000  & Pairwise & \textbf{7.4M}  & 86.7 & \textbf{89.8} & \textbf{99.7} & 99.2 \\
MLP 1 x 10000 & k-WTA FC & 7.9M           & 84.7 & 82.7          & \textbf{99.7} & 98.7 \\ \hline
MLP 3 x 700   & Pairwise & 2.8M           & 73.0 & \textbf{77.1} & \textbf{99.2} & 98.3 \\
MLP 3 x 1000  & k-WTA FC & 2.8M           & 63.3 & 59.8          & 99.1          & 97.7 \\ \hline
CNN 1 layer   & Pairwise & 0.1M           & 80.9 & \textbf{81.0} & \textbf{99.6} & 98.2 \\
              & k-WTA FC & \textbf{0.03M} & 54.3 & 58.2          & 99.3          & 98.3 \\ \hline
CNN 2 layers  & Pairwise & 0.5M           & 78.6 & \textbf{80.5} & \textbf{99.7} & 98.6 \\
              & k-WTA FC & \textbf{0.2M}  & 58.4 & 57.2          & 99.3          & 98.0  
\end{tabular}
\end{table}
Table \ref{tab:split-mnist} shows our results on single-head and multi-head Split MNIST experiments. We focus more on the single-head results because we feel like that is the more interesting and realistic problem. For the multi-head, we will simply state that most of our methods get over 99\% accuracy, as do many others in the literature, with the best result we could find being 99.97\% with ModelZoo \citep{model_zoo_ramesh2022model}. However, we believe ours to be the first method to reach 99\% accuracy when trained in an online fashion and only for a single epoch. S-MAS underperforms slightly in the multi-head results, but it might be because it's not very well tuned to the task. Networks with the pairwise output layer generally perform a little bit better.

Moving on to the single-head results, we see more interesting variety in the results. The difference between pairwise and fully connected output heads is much more pronounced, with pairwise outperforming FC by at least 5\% and sometimes over 20\%. The roles of S-MAS and Adagrad are switched, with S-MAS outperforming Adagrad in almost all cases.

Another interesting trend is that deeper backbone architectures (MLP 3x and the 2 layer CNN) are significantly worse than smaller and shallower networks. This has been observed before, and for example, \citet{lan2023elephant} and \citet{bricken2023sparse} give results only for networks that have 1 or 2 layers being trained. However, we note that in the case of a 3-layer MLP, the performance degrades less with the pairwise output head compared to an FC output head.

The best results we found from the literature for rehearsal-free, single-head 5 task Split MNIST are from \citet{bricken2023sparse}, with SDMLP+EWC reaching 83\% accuracy with 0.8M parameters and FlyModel \citep{shen2021algorithmic} reaching 91\% accuracy with 10k hidden neurons. PEC \citep{zajac2024prediction} reached 92.3\% with a generative method. In the 0.8M parameter case, we outperform SDMLP+EWC with our pairwise architecture reaching \textbf{84.4\%} accuracy despite SDMLP+EWC using the task boundaries for regularization and training for 500 epochs. Our larger models do not reach the accuracy of the FlyModels, which is an associative rule learning system. But with \textbf{89.8\%} accuracy, it does beat the best discriminative gradient descent-based method, which is again SDMLP+EWC with the accuracy of 86\%.

We also got an accuracy of \textbf{81.0\%} with a small 0.1M parameter CNN network with a pairwise output layer. The most directly comparable result is 73.2\% with ECNN \citep{lan2023elephant}.

\subsection{Performance on Split Fashion-MNIST}

\begin{table}[h]
\centering
\caption{Split Fashion-MNIST}
\label{tab:split-fashion-mnist}
\begin{tabular}{ccc|cc|cc}
Network &
Output &
Total &
Single-head &
Single-head &
Multi-head &
Multi-head \\ 
backbone & head type & parameters & Adagrad & S-MAS & Adagrad & S-MAS
\\ \hline
MLP 1 x 700 & Pairwise & 0.8M           & 65.0 & \textbf{69.1} & \textbf{99.0} & 94.5 \\
MLP 1 x 1000 & k-WTA FC & 0.8M                & 64.2 & 64.1 & 98.9 & 97.9 \\ \hline
MLP 1 x 3000 & Pairwise & \textbf{7.4M} & 71.2 & \textbf{72.6} & \textbf{99.1} & 98.7 \\
MLP 1 x 10000 & k-WTA FC & 7.9M               & 68.8 & 69.4 & \textbf{99.1} & 98.5 \\ \hline
MLP 3 x 700 & Pairwise & 2.8M           & 63.4 & \textbf{67.2} & \textbf{98.9} & 98.2 \\
MLP 3 x 1000 & k-WTA FC & 2.8M                & 60.2 & 58.8 & 98.6 & 97.4 \\ \hline
CNN 1 layer & Pairwise & 0.1M           & 57.2 & \textbf{68.0} & \textbf{99.1} & 97.8 \\
& k-WTA FC & \textbf{0.03M}                   & 55.0 & 58.9 & 98.9 & 98.3 \\ \hline
CNN 2 layers & Pairwise & 0.5M          & 61.5 & \textbf{63.9} & \textbf{98.9} & 98.1 \\
& k-WTA FC & \textbf{0.2M}                    & 55.3 & 55.3 & 98.7 & 97.8
\end{tabular}
\end{table}

Split Fashion-MNIST is a slightly harder continual learning experiment than Split MNIST. It is also relatively niche, so we will keep the analysis of the results short. The multi-head version of the experiment is still relatively easy for our architectures, but the single-head performance is clearly worse than in Split MNIST.

The best results for single-head Split Fashion-MNIST are again from \citet{bricken2023sparse}, with SDMLP+EWC reaching 74\% with 0.8M parameters and FlyModel getting 76\% with 10k hidden neurons. We fall a little short of both of these with \textbf{69.1\%} and \textbf{72.6\%} respectively. However, the same caveats from \ref{sec:split-mnist} still apply to SDMLP+EWC and FlyModel results. \citet{bricken2023sparse} also reported 73\% accuracy with the plain SDMLP, but we believe this to be a clerical error since we got 64\% accuracy when running the official code on this experiment, which would be better in line with the other results.

\subsection{Performance on Permuted MNIST}

\noindent 
\begin{minipage}[t]{0.33\textwidth} 
Permuted MNIST is another popular continual learning variant of the MNIST dataset. In permuted MNIST, each task in the sequence of tasks is created by applying a new fixed permutation to the pixels of the original MNIST images.

We did the 10 permutation version of this experiment and report the mean performance over 10 random network initialization, and used different random permutations for each run.

\end{minipage} \hfill
\begin{minipage}[t]{0.65\textwidth} 
\vspace{-1cm}
\begin{table}[H]
\centering
\caption{Permuted MNIST overall mean validation accuracy}
\label{tab:permuted-mnist}
\begin{tabular}{ccc|cc}
Network  & Output    & Total      &  &  \\
backbone & head type & parameters & Adagrad & S-MAS \\ \hline
MLP 1 x 700   & Pairwise & 0.8M & \textbf{95.4} & 92.9 \\
MLP 1 x 1000  & k-WTA FC & 0.8M & 95.3 & 90.9 \\ \hline
MLP 1 x 3000  & Pairwise & \textbf{7.4M} & \textbf{97.3} & 95.7 \\
MLP 1 x 5000  & k-WTA FC & 7.9M & 96.2 & 92.3 \\ \hline
MLP 3 x 700   & Pairwise & 2.8M & \textbf{88.3} & 87.7 \\
MLP 3 x 1000  & k-WTA FC & 2.8M & 84.7 & 81.6 \\
\end{tabular}
\end{table}
\end{minipage}

HAT \citep{hat_serrà2018overcoming} is a strong baseline on Permuted MNIST. They got 97.4\% accuracy with 0.7M parameters and 98.6\% with 5.8M parameters. However, both of these are two layers deep MLP, and it's unclear if HAT could utilize the parameters more efficiently in a single-layer configuration. Table \ref{tab:permuted-mnist} shows our results that are a little worse than HAT, with \textbf{95.4\%} accuracy using 0.8M parameters and \textbf{97.3\%} using 7.4M parameters. Pairwise and FC are about equal with 0.8M parameters, but pairwise scales better to larger models. We believe our results are the best by any task-agnostic method, but we did not find many good comparisons. \citet{Iyer_2022} reported 94.6\% and \citet{zeno2019task} ~94\%, but both of these methods try to infer task boundaries, so in our classification they are task-free but not task agnostic.

\begin{figure}[h]
\begin{center}
\includegraphics[scale=0.5]{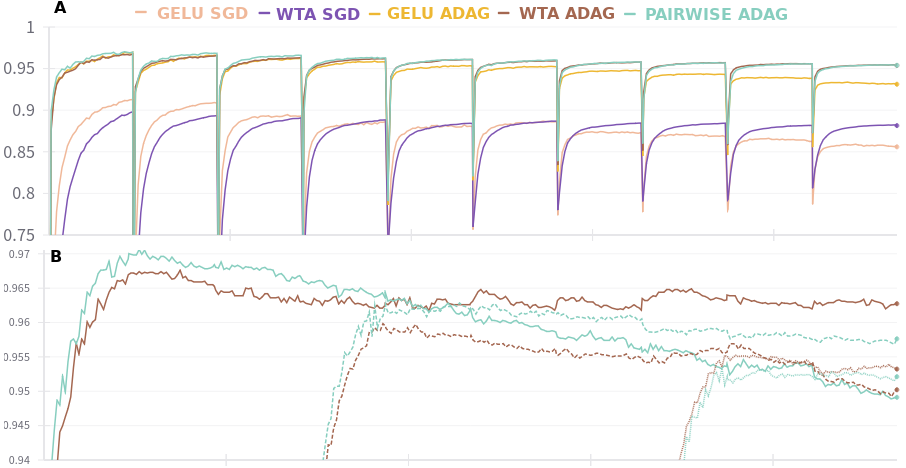}
\end{center}
\caption{(A) Overall accuracy on all the tasks learned so far in Permuted MNIST with the small MLP architectures (0.8M parameters). WTA Adagrad and Pairwise Adagrad accuracies overlap almost exactly for most of the training. (B) Accuracies on the first, fourth, and eighth tasks for WTA Adagrad and Pairwise Adagrad.}
\label{fig:pminst-perf}
\end{figure}

Figure \ref{fig:pminst-perf} shows the overall accuracies in a single training run for 5 different methods with 0.8M parameter MLP architectures. GELU-based methods are clearly worse at retaining old task performance. SGD needs a lower learning rate to reach the best final performance and is already behind Adagrad during the first task. Pairwise and plain FC WTA are about equal in overall performance, but \ref{fig:pminst-perf}B shows that they can actually have different performances in the individual tasks.

\section{Ablation studies and other results}

\subsection{Analysis of Network Sparsity}

To test how the sparsity/density hyperparameter of the k-WTA activation function affects the results, we did a small test on single-head Split MNIST while varying the amount of sparsity.

Because we ran our experiments with many different architectures with different hidden layer widths, we give the desired density of activations as a percentage instead of a number. For example, WTA with density $p = 10\%$ and hidden dimension of 3000 is equal to k-WTA with $k = 300$.

\begin{figure}[h]
\begin{center}
\includegraphics[scale=0.25]{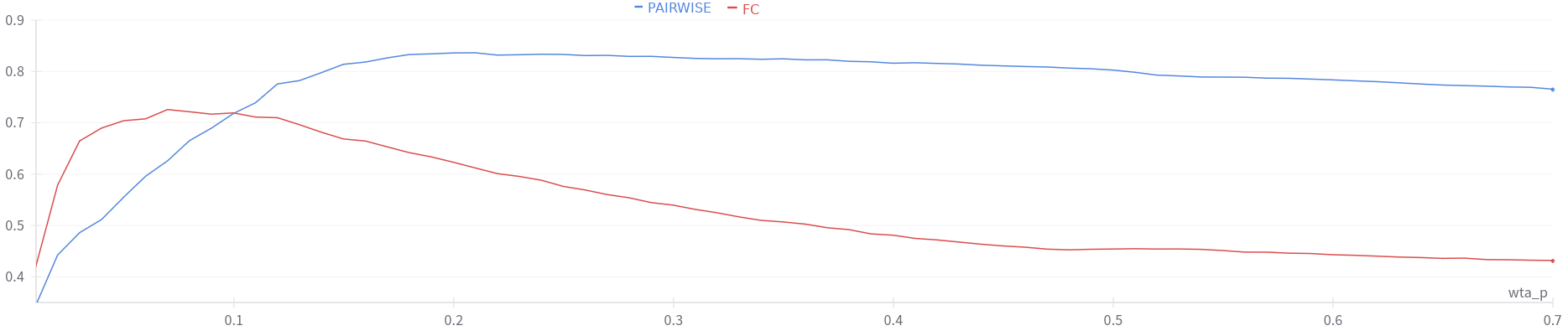}
\end{center}
\caption{Split MNIST with the small MLP architecture (0.8M parameters) with different values for hidden layer sparsity.}
\label{fig:sparsity}
\end{figure}

Figure \ref{fig:sparsity} shows how Pairwise and FC output heads work with different levels of representation density. In general, the fully connected output has a rather narrow zone where it works well at around 8 to 10\% density. The pairwise layer needs a little more density for optimal performance and peaks around 20\% density but suffers a lot less from too much density. Intuitively, it makes sense to us that the pairwise layer works better with higher density since the feature crossing expansion exaggerates the effects of sparsity. Still, it is surprising that the pairwise layer can maintain over 75\% final accuracy all the way to 70\% density of activations.

\subsection{Performance with SGD}
\label{sec:sgd}

\noindent 
\begin{minipage}[t]{0.4\textwidth} 

We present the evaluation of our models using vanilla Stochastic Gradient Descent on the Split MNIST dataset. This is to evaluate the models' performance under a basic optimization algorithm. Continual learning with basic SGD is considerably more challenging because the learning rate stays constant throughout the training and same for each model weight, so even weights that are very important for previous tasks can be quickly overwritten.




\end{minipage} \hfill
\begin{minipage}[t]{0.55\textwidth} 
\vspace{-1cm}
\begin{table}[H]
\centering
\caption{Split MNIST with SGD. Mean validation accuracy of 30 runs with different random seeds.}
\label{tab:sgd}
\begin{tabular}{ll|cc}
Backbone     & Output   & \multicolumn{1}{l}{Single-head} & \multicolumn{1}{l}{Multi-head}   \\
Architecture & Head     & \multicolumn{1}{l}{Split MNIST} & \multicolumn{1}{l}{Split MNIST}  \\ \hline
MLP 1 x 700  & Pairwise & \textbf{69.0}                   & \textbf{98.3}                    \\
MLP 1 x 1000 & k-WTA FC & 51.1                            & 97.7                             \\ \hline
CNN 1 layer  & Pairwise & \textbf{62.0}                   & \textbf{98.6}                    \\
CNN 1 layer  & k-WTA FC & 39.9                            & 98.5                            
\end{tabular}
\end{table}
\end{minipage}

Table \ref{tab:sgd} shows the results on Split MNIST with SGD. The relative order of architectures stayed the same between these results and the ones in \ref{sec:split-mnist}, with each architecture performing roughly 15-20 percentage points worse than with S-MAS. The small 0.8M parameter MLP with Pairwise output head got to \textbf{69.0\%} accuracy, which is still better than similarly sized ReLU models get with EWC (61\%), SI (36\%) or MAS (49\%) \citep{bricken2023sparse}. Even the fully connected network with WTA activation scores higher than SI and MAS at 51.1\% accuracy. These results suggest again that the network architecture is more important for continual learning performance than the choice of an algorithm.

Figure \ref{fig:pminst-perf} has more comparisons with SGD and different activation functions in the Permuted MNIST experiment.

\subsection{Effect of batch size}
\label{sec:batch_size}
\noindent 
\begin{minipage}[t]{0.63\textwidth} 
One of the goals of online learning is to have the ability to use no mini-batches or at least very small mini-batches. Therefore, we tested how our system works with smaller batch sizes. Table \ref{tab:batch_size} shows our best results on Split MNIST with a batch size of 16 and 1 using the 0.8M parameter MLP with a pairwise output head. Overall, the accuracy drops less than 2\%, going from a batch size of 64 down to 1.

One thing to note is that in S-MAS, the parameter $\lambda$ that controls the stability-plasticity trade-off depends on the batch size and needs to be changed accordingly. This is because, with a smaller batch size, there are more importance updates, and a high $\lambda$ makes the network too stable too fast.
\end{minipage} \hfill 
\begin{minipage}[t]{0.33\textwidth} 
\vspace{-1cm}
\begin{table}[H]
\centering
\caption{Single-head Split MNIST with S-MAS while varying the batch size. The result with bs 64 is the mean of 30 random seeds, while bs 16 and 1 are the results of a single run.}
\label{tab:batch_size}
\begin{tabular}{ccc}
Batch size & $\lambda$ & Mean accuracy  \\   \hline
64  &  0.1  & 84.4 \\
16  & 0.01  & 83.4 \\
1   & 0.005 & 82.9 \\
\end{tabular}
\end{table}
\end{minipage}

With Adagrad, the best result we got on the single-head Split MNIST experiment with a batch size of 1 was \textbf{79.2\%}.

\section{Conclusion}

In this paper, we introduced a new type of continual learning architecture and validated its merits in online task-agnostic experiments. The pairwise layer shows promising results in our experiments and, on some benchmarks, beats the best comparable results found in the literature. We expect that engineered features, like the pairwise feature crosses, will be generally useful in continual learning and that choosing the network architecture should be a key consideration when designing a continual learning system.

This study was conducted to deepen the knowledge in the field of rehearsal-free, online continual learning. We do not see that the study had any wider societal impacts. All the experiments were run on a single consumer-grade GPU (RTX 3090).

\subsection{Future Directions}
We implement parameter sparsity in PW-layer to make it cheaper to compute and reduce the number of trainable parameters. In this study, we always randomly initialize the sparseness and keep it static through the training. However, this random initialization is likely far from the optimal arrangement. It would be interesting to try to adapt the sparse mapping between cross-features and outputs to fit the training data. Think of how, in brains, useless synapses die and are replaced by useful synapses over time. 
AutoCross \citep{luo2019autocross} is an algorithm for finding better cross features in tabular data. AutoCross also tries to find cross features with more than just two inputs.

Another potential direction is looking for better task-agnostic continual learning algorithms. We feel this search is partially motivated by the unreasonable effectiveness of Adagrad. Perhaps one option is to fit existing regularization algorithms like EWC or SI into our streaming parameter importance calculation framework.


\bibliography{collas2024_conference}
\bibliographystyle{collas2024_conference}

\appendix
\section{Code for reproducing}

The code supporting the findings of this study is available online for review and replication purposes. To ensure the reproducibility of our results, we have made the source code publicly accessible. 

The code repository, including documentation and setup instructions, can be accessed  \href{https://github.com/skeskinen/pairwise_online_learning}{here}.

We encourage researchers and practitioners to check, utilize, and extend our work.

We also include with the code a directory of .json config files with the hyperparameters for almost all of the experiment results found in this study. This should make it extra easy to reproduce any specific result.

\section{Network initialization}

We used Kaiming He initialization \citep{he2015delving} for all the backbones and the fully connected output layers.

For the Pairwise layers, we tried a bunch of different adapting initialization methods, including: He normal, He uniform, Xavier normal, Xavier uniform, etc. The best one out of the standard ones was LeCun Normal, but even better was simple normal initialization with a constant $std = 0.001$, which we used for all of the experiments. 

\section{Preliminary testing on Split CIFAR-10}
\label{sec:cifar}

Having achieved some success on MNIST and FashionMNIST, it would be nice to take the next step and move to bigger datasets. To this end, we did some preliminary testing on 5 task split CIFAR-10. Again, we focus more on the single-head case since we feel it is more realistic for future applications. However, this experiment already seems to be too challenging, and the results are quite poor. The best results were around \textbf{31\%}, which is just 11 percentage points better than only learning the last task perfectly. While this result is still better than any we could find for single-head, non-rehearsal, non-pretrained split CIFAR-10 (e.g. \citet{lan2023elephant} reported 24.3\% accuracy), we feel that 31\% accuracy still amounts to essentially failing the experiment. Therefore, we decided to focus more on the easier experiments for now and did not do a full experiment on CIFAR-10. Realistically, 31\% is not an awful result considering that \citet{aljundi2019gradient} reported under 30\% performance with iCarl \citep{rebuffi2017icarl} and GEM \citep{lopezpaz2022gradient} which are rehearsal methods. \citet{ye2022taskfree} reached 52.7\% with another rehearsal method.

With multi-head setup, the best preliminary results we got on 5 task split CIFAR-10 were around \textbf{84\%} with 1 layer CNN backbone and only a negligible difference between Pairwise and WTA FC output heads. Random guessing would give an accuracy of 50\%.

\end{document}

%% file: math_commands.tex
\usepackage{amsmath,amsfonts,bm}

\def\eqref#1{equation~\ref{#1}}

\def\1{\bm{1}}

\DeclareMathAlphabet{\mathsfit}{\encodingdefault}{\sfdefault}{m}{sl}
\SetMathAlphabet{\mathsfit}{bold}{\encodingdefault}{\sfdefault}{bx}{n}

